\documentclass{article}
\usepackage{PRIMEarxiv}

\usepackage[utf8]{inputenc} 
\usepackage[T1]{fontenc}    
\usepackage{hyperref}       
\usepackage{url}            
\usepackage{booktabs}       
\usepackage{amsfonts}       
\usepackage{nicefrac}       
\usepackage{microtype}      
\usepackage{lipsum}
\usepackage{fancyhdr}       
\usepackage{graphicx}       
\graphicspath{{media/}}     
\usepackage{amsmath}
\usepackage{microtype}      
\usepackage{cleveref}       
\usepackage{lipsum}         
\usepackage{graphicx}
\usepackage[sort&compress,numbers]{natbib}
\usepackage{doi}
\usepackage{enumerate}
\usepackage{amssymb,enumitem}
\usepackage{multirow}
\usepackage{xcolor}
\usepackage{longtable}
\usepackage{algorithm}
\usepackage{algpseudocode}
\usepackage{array}
\makeatletter
\def\algbackskip{\hskip-\ALG@thistlm}

\usepackage{subcaption}
\usepackage{listings}

\title{Application of Self-Supervised Learning to MICA Model for Reconstructing Imperfect 3D Facial Structures}
\author{
Phuong D. Nguyen$^{1,}$, Thinh D. Le$^{1,}$, Duong Q. Nguyen$^{2}$,  Binh Nguyen$^{3}$, H. Nguyen-Xuan$^{1,}$\thanks{Corresponding author: CIRTECH Institute, HUTECH University (Email: ngx.hung@hutech.edu.vn).}  \\
$^{1}$CIRTECH Institute, HUTECH University, Ho Chi Minh City, Viet Nam \\
$^{2}$Department of Mathematics and Statistics, Quy Nhon University, Quy Nhon City, Viet Nam\\
$^{3}$Faculty of Mathematics and Computer Science, HCMVNU-University of Science, Ho Chi Minh City, Viet Nam\\
}
\begin{document}
\maketitle
\begin{abstract}
In this study, we emphasize the integration of a pre-trained MICA model with an imperfect face dataset, employing a self-supervised learning approach. We present an innovative method for regenerating flawed facial structures, yielding 3D printable outputs that effectively support physicians in their patient treatment process. Our results highlight the model's capacity for concealing scars and achieving comprehensive facial reconstructions without discernible scarring. By capitalizing on pre-trained models and necessitating only a few hours of supplementary training, our methodology adeptly devises an optimal model for reconstructing damaged and imperfect facial features. Harnessing contemporary 3D printing technology, we institute a standardized protocol for fabricating realistic, camouflaging mask models for patients in a laboratory environment.
\end{abstract}

\keywords{3D printing technology \and face reconstruction \and self-supervised learning \and 3D printed model}

\section{Introduction}
\begin{figure}[!h]
\centering
\includegraphics[scale=0.8]{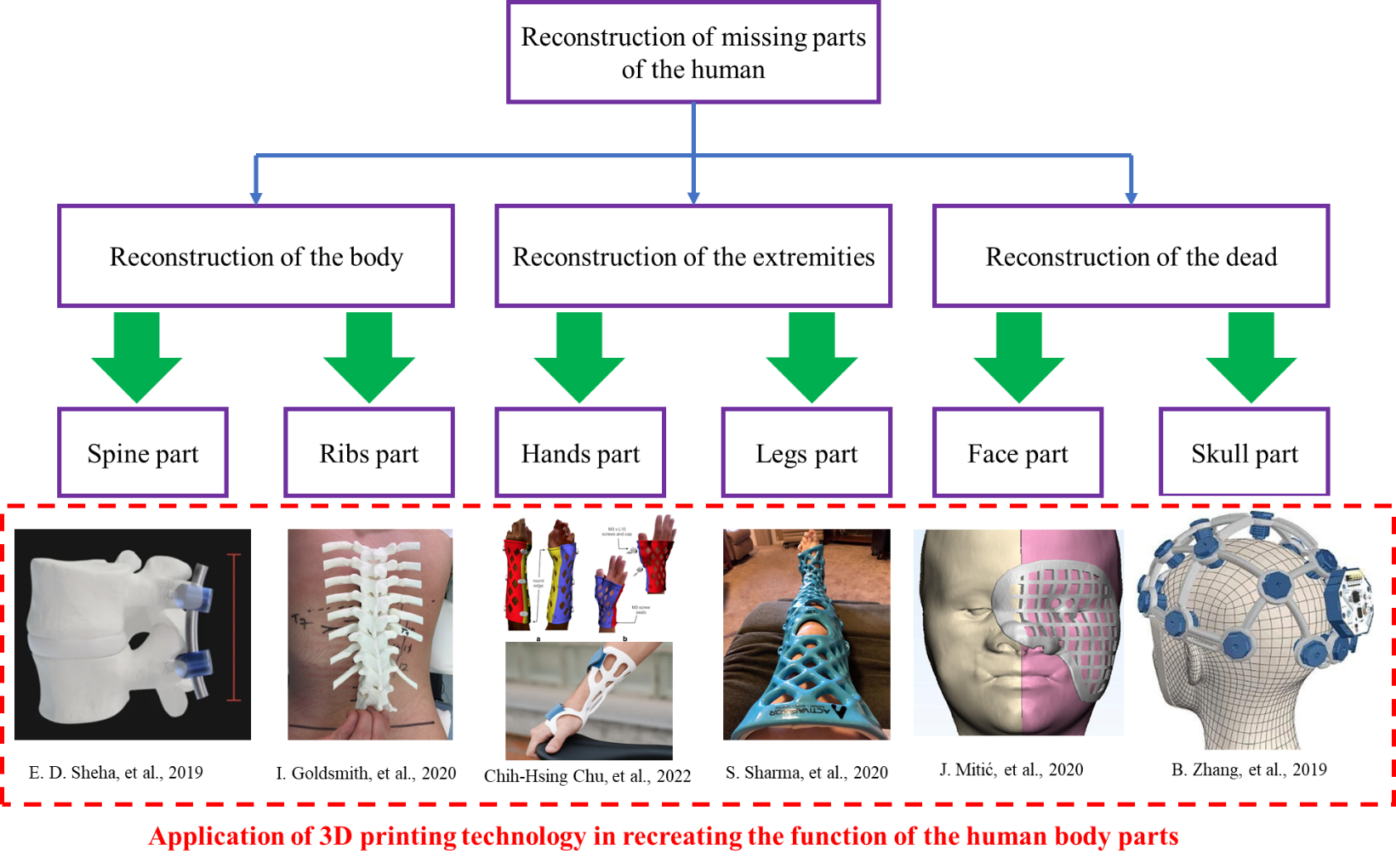}
\caption{Replacing some parts of the human body with 3D printing technology}
\label{F1}
\end{figure}

Currently, people who are injured due to traffic accidents, occupational accidents, birth defects, diseases have made them lose a part of their body. As illustrated, many parts of the human body are injured and lose their natural function as shown in Fig. \ref{F1}. Herein, the head area, which is soft and  most vulnerable to external impacts, has the highest injury rate and probability of death. Reconstructing the wound and creating a product to cover it is crucial for patients to reduce the risk of re-injury, improve aesthetics, and increase confidence \cite{ meisel_digital_2022, asanovic_development_2019}. This research introduces an alternative way to reduce the risk of re-injury and avoid impact when operating. It also helps to increase aesthetics and confidence for people with injured faces. In addition, with the advancements in 3D printing and scanning technologies in rapid prototyping, the reconstruction process of organs  has become faster, more accurate, and safer \cite{li_design_2020, cheng_topological_2020,   popescu_fast_2020}. However, recreating the functions and the true shapes of upper facial features, which have been damaged by injury, remains highly challenging and requires multiple trials in the treatment process.

\begin{figure}[!h]
\centering
\includegraphics[scale=0.55]{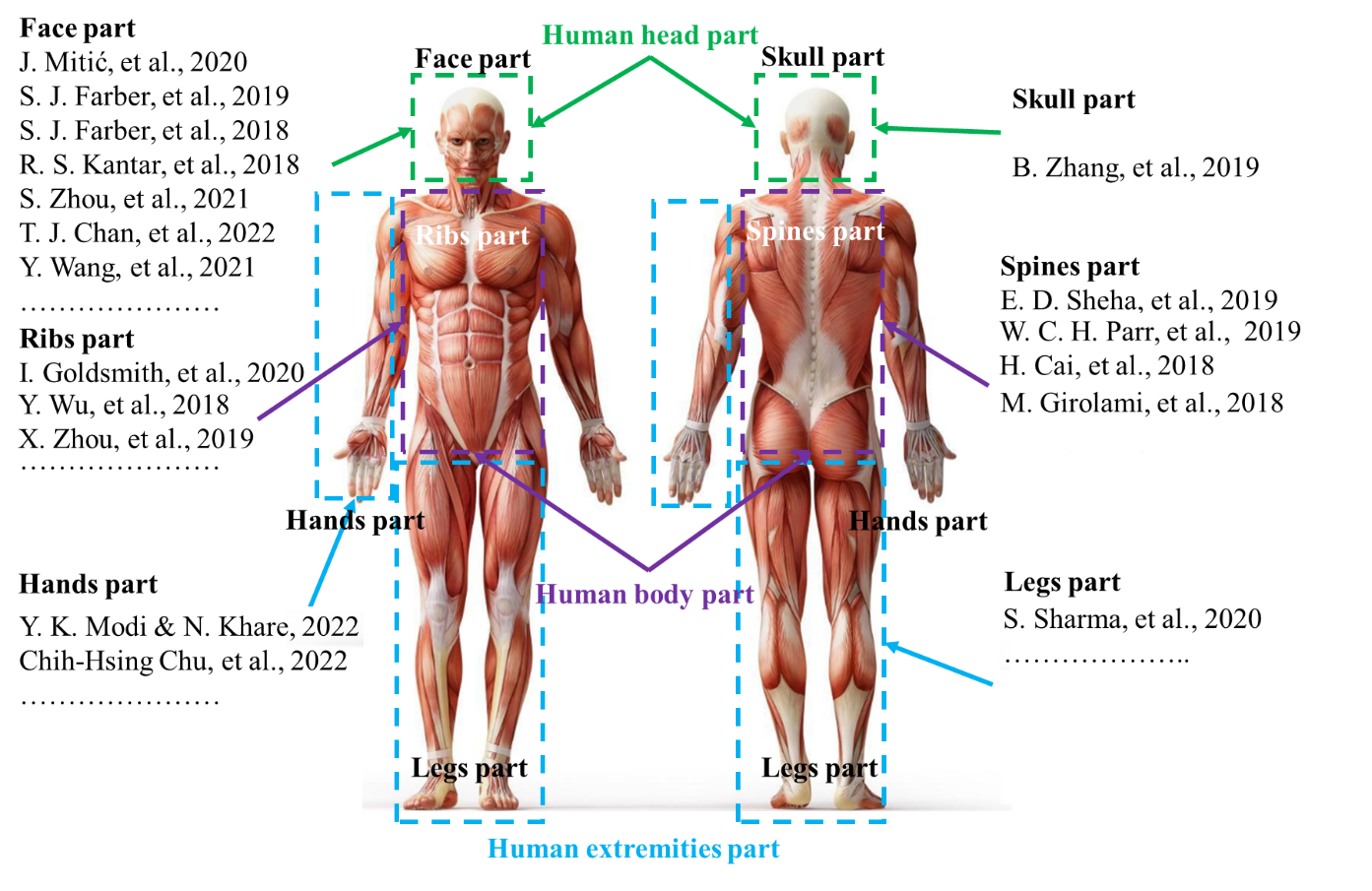}
\caption{Some studies on reconstruction of parts of the human body}
\label{some_studies_on_reconstruction_of_parts_of_the_human_body}
\end{figure}

The reconstruction of human body parts is divided into three main research directions including \cite{  yin_tailored_2022}: the reconstruction of the body \cite{ira_goldsmith_chest_2022, alvarez_design_2021}, the reconstruction of the extremities \cite{poier_development_2021, li_feasibility_2018}, and the reconstruction of the head, as shown in Fig. \ref{F1}. The reconstruction of extremities has  garnered significant research interest as compared  to other areas of the human body, including the reconstruction of bones in the hands \cite{sheha_3d_2019,modi_patient-specific_2022, chih-hsing_chu_customized_2022} and the legs \cite{ sharma_utilization_2020}. The reconstruction of the body is often divided into two parts: the reconstruction of the spine \cite{  girolami_biomimetic_2018,  parr_3d_2019, cai_3d_2018} and the reconstruction of the ribs \cite{goldsmith_chest_2020,  wu_application_2018, zhou_analysis_2019}. Previous studies have been focusing on the reconstruction of the body and the extremities as shown in Fig. \ref{some_studies_on_reconstruction_of_parts_of_the_human_body} because the main reason is that these areas are rather less  susceptible to injury compared to other areas of the human body and they are also relatively easy to implement, easy to assemble, and less risky for patients in the treatment process. The upper body can be divided into two distinct research groups, namely, the reconstruction of the face \cite{mitic_reconstruction_2020, farber_reconstructing_2019} and the reconstruction of the skull \cite{ zhang_3d_2019}. Due to the high fatality rate associated with injuries sustained in this region, it is considered the most perilous area for humans. Consequently, research on this region of the human body is relatively scarce compared to other areas.

Having analyzed the unique risks associated with the patient’s head area, we recognize the importance of conducting thorough studies to enhance patient safety and optimize outcomes in this critical area.  A study on patients with head trauma treated at the Walter Reed National Military Medical Center (WRNMMC) was conducted, as described by Farber et al. \cite{farber_reconstructing_2019, farber_face_2018, kantar_single-stage_2018}. The application of microsurgery techniques to treat severe head wounds in a war zone was implemented for the first time. This study explored various methods to heal these injuries with the aim of achieving the best possible outcome for patients. In addition, Shuang-Bai Zhou and colleagues \cite{zhou_strategy_2021, chan_state_2022} proposed a method to reconstruct the deformities of the nose or the central points of the face using forehead-expanded flaps. The method was performed on 22 patients, including 13 men and 9 women. The outcome was that 17 people had burns and 5 people had other injuries. This method of using forehead-expanded flaps was shown to be capable of  reconstructing the deformities of the nose and mouth in patients with only one extensive treatment. This flap can be flexibly adjusted to fit the size of the deformity right in the middle of the face. Wang and his colleagues \cite{wang_aesthetical_2021} used a combination of virtual surgical planning and 3D printing in reconstructing facial deformities on the jawbone. Study data were collected from the jaw structure of patients in the time period from 2013 to 2020. The patients were divided into two groups, in which group 1 (20 people) used the above technique and group 2 (14 people) used conventional manual surgery. The results indicate that the present approach offers numerous advantages to patients receiving the treatment. The positive outcomes of the aforementioned studies have established a new trend in providing care for those who have been injured in hazardous areas. The proposed method generates great potential for regenerating and rejuvenating various body parts. However, it is important to note that the study  is only limited to an experimental setting and has not been implemented in a practical manner yet. 
\begin{figure}[!h]
\centering
\includegraphics[scale=0.65]{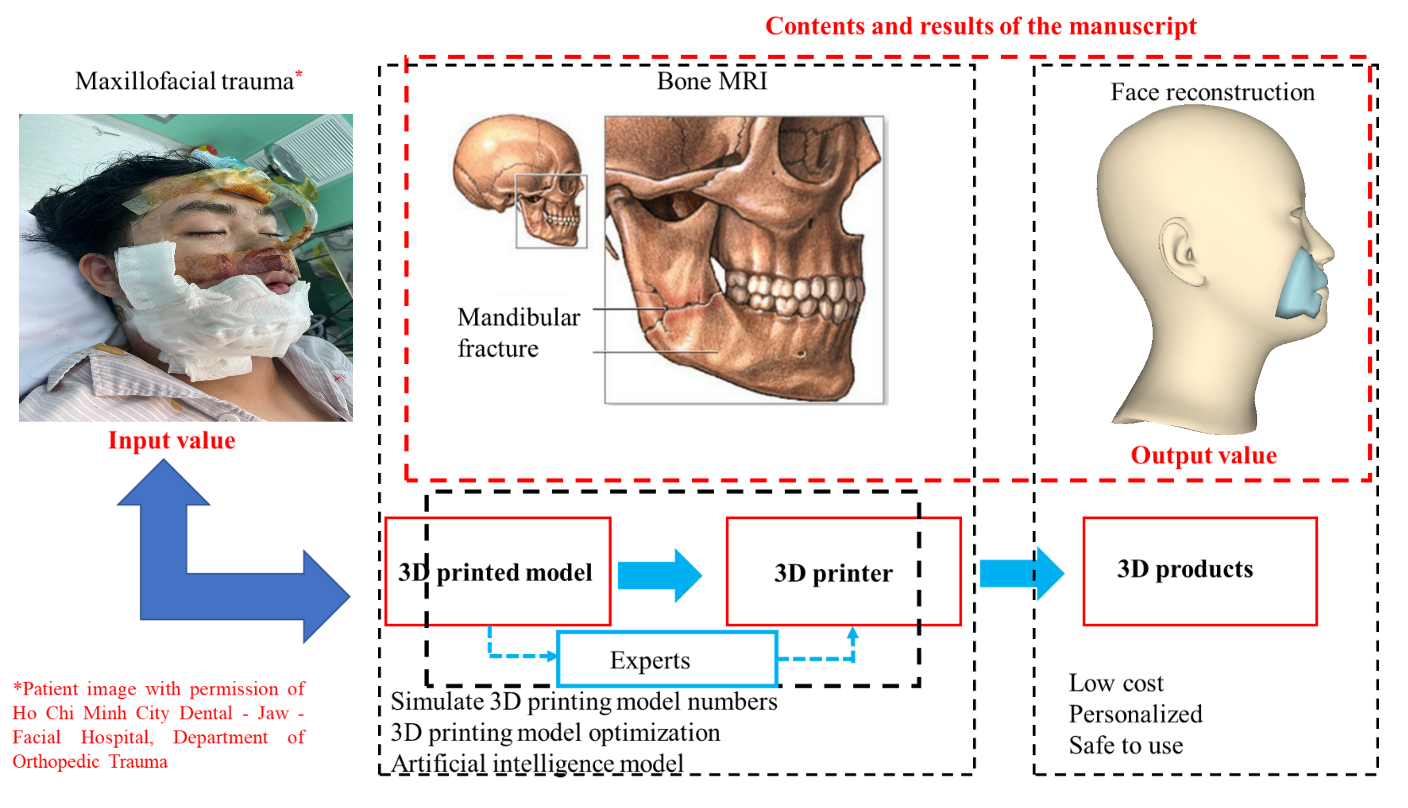}
\caption{Proposed model and results from this study}
\label{F3}
\end{figure}

In our previously published study \cite{nguyen_3d_2023}, we demonstrated promising results in 3D reconstructing imperfect faces. We identified the limitations of current methods for creating new body parts to replace damaged ones and proposed an effective solution. Specifically, we implemented a process using 3D printing technology and deep learning networks to extract the filling to aid in masking for patients with imperfect facial defects, a novel approach known as a wound covering, as depicted in Fig. \ref{F3}. This approach helps patients with facial trauma regain their confidence in daily life while also protecting and cleaning the wound from external elements. Additionally, we introduced a dataset capable of covering multiple facial trauma cases. However, the deep learning model used in the study required significant computational resources for training due to its complexity, resulting in training times of several days, presenting a challenge to conducting research in this field.

In this study, we aim to address these limitations by achieving the following objectives:
\begin{itemize}
\item Utilize the MICA model \cite{zielonka_towards_2022} to reconstruct imperfect faces in 3D format with shorter training times, taking only a few hours,
\item Proposing a new training strategy by combining pre-train MICA model and self-supervised learning to enhance its performance in this specific application,
\item Utilize the same dataset to compare the effectiveness of the two methods in addressing the challenges of extracting the filling to aid in masking for patients with facial trauma.
\end{itemize}

\section{Proposed approach}
\begin{figure}[!h]
\centering
\includegraphics[scale=0.2]{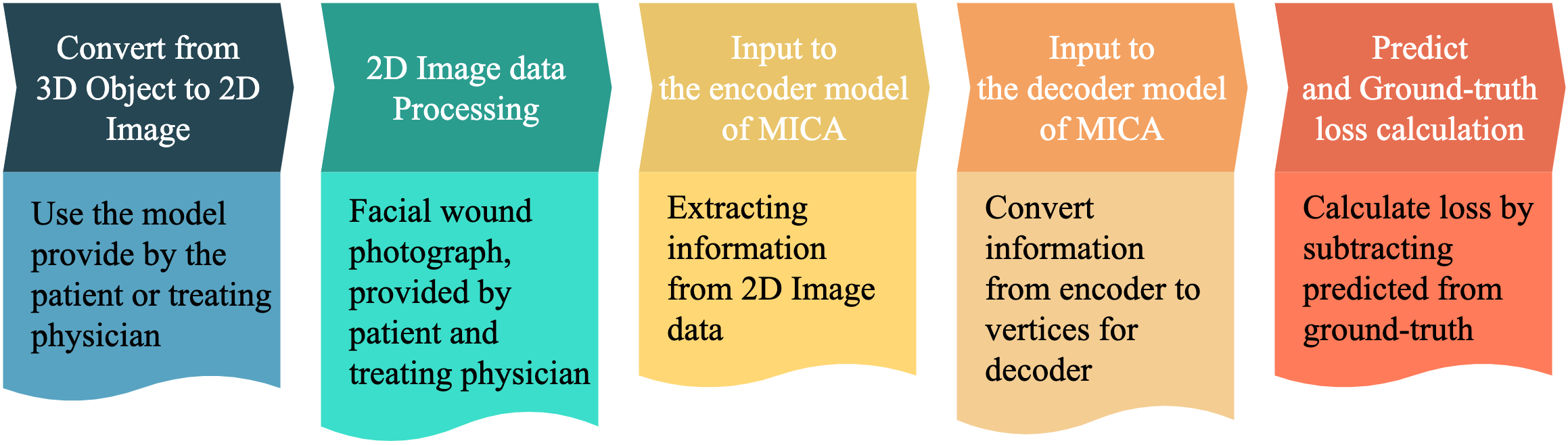}
\caption{The whole process of the algorithms }
\label{the_whole_process_of_the_algorithms}
\end{figure}

\begin{figure}[!h]
\centering
 \includegraphics[scale=0.4]{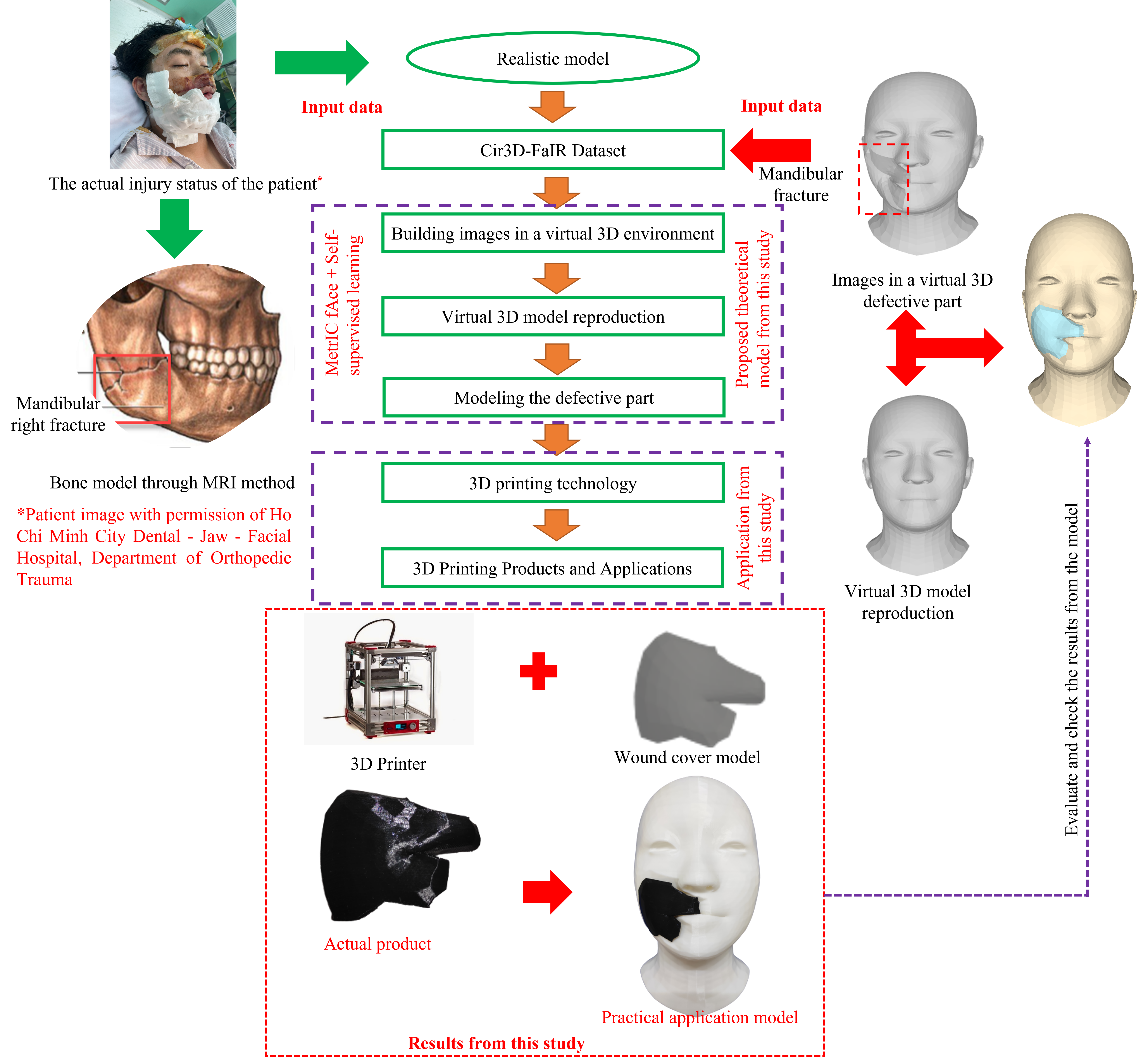}
\caption{Our proposed approach}
\label{F5}
\end{figure}

\begin{figure}[!h]
\centering
\includegraphics[scale=0.2]{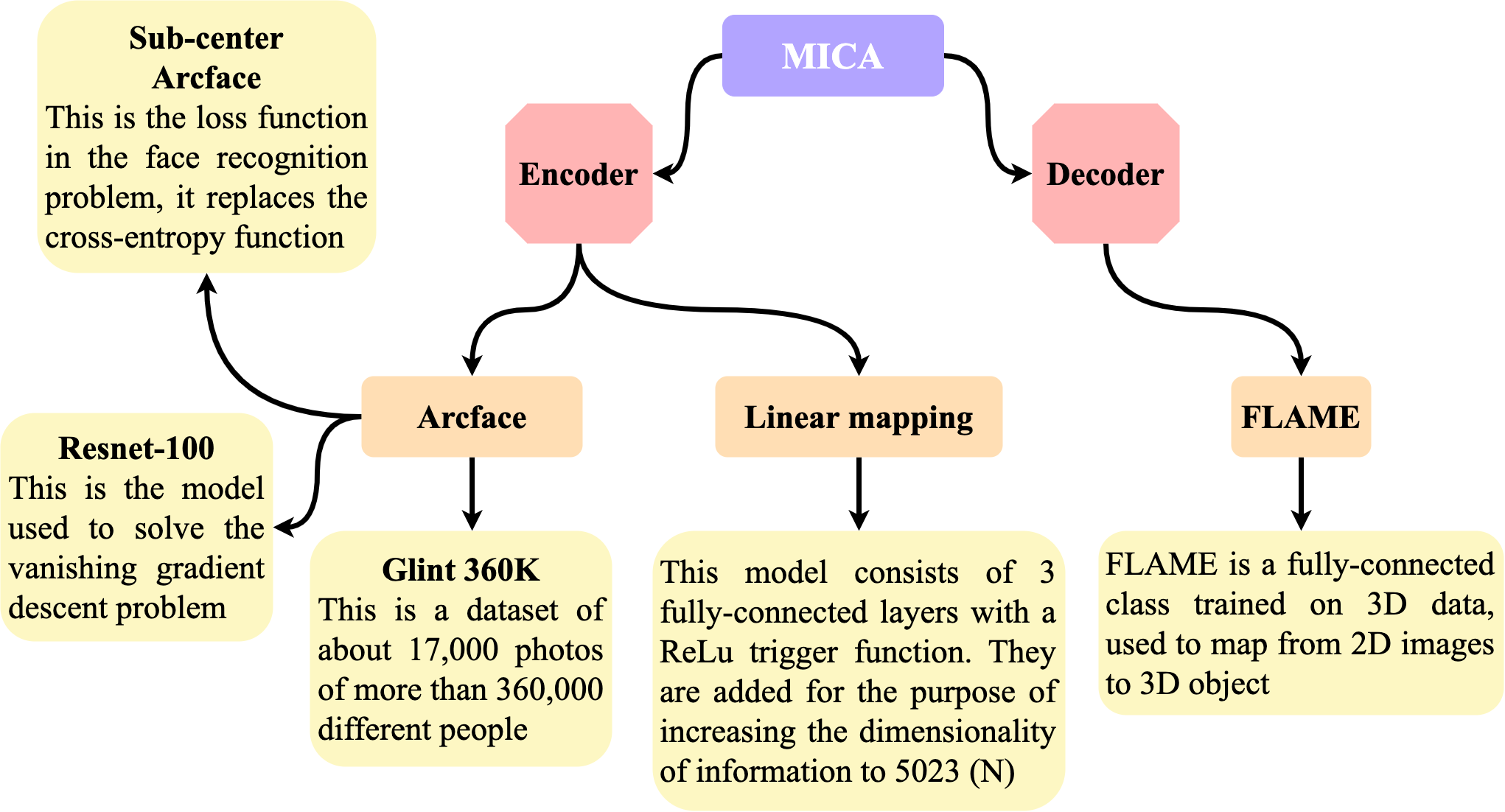}
\caption{A framework of the MICA model}
\label{MICA}
\end{figure}

In this study, we solve the problem posed based on the MICA model \cite{zielonka_towards_2022}, the implementation process is shown in Fig. \ref{the_whole_process_of_the_algorithms}.
Among 3D face reconstruction models, the MICA model stands out as effective due to its ability to analyze facial features and reconstruct finer details compared to previous methods. Additionally, it performs well on evaluation datasets with high reliability. In this study, the MICA model is applied with a 2D image input, which is highly suitable as it requires fewer computing resources and is more cost-effective than 3D input formats, with training time reduced from several days to several hours. Therefore, we employed two pre-training techniques for the MICA model: transfer learning and self-supervised learning, to find the best approach for this specific study. We  modify the inputs and outputs in the process of reconstructing the patient's injured head area as shown in Fig. \ref{F5}.  The input is a 2D image, which shows the actual patient's face with the wound. The output is a 3D mesh restored to its pre-injury state. Thus, the MICA model includes two main components: Identity Encoder and Geometry Decoder as shown in Fig. \ref{MICA}.

\subsection{Identity Encoder}
Identity Encoder is formed by two main components: ArcFace model and Linear Mapping model. The goal of Encoder is to extract identity-specific information from an input 2D image using two components: ArcFace and Linear Mapping. First, the ArcFace model determines the level of similarity  between the input image and the reference image of the same identity, and then generates a feature vector  to capture identity-specific information. Second, the Linear Mapping model maps this feature vector to a higher dimensional space to improve the discriminative power of the feature representation.
\begin{itemize}
\item ArcFace: The ArcFace architecture \cite{deng_sub-center_2020, zhao_towards_2020, deng_arcface_2019}- a model developed based on ResNet-100 - uses a shortcut to map input from previous layers to following layers, thereby avoiding the phenomenon of vanishing gradients \cite{he_deep_2016, roodschild_new_2020, hochreiter_recurrent_1998}. This is important for updating model weights and enabling the model to learn the best features. At the same time, the additive angular margin loss function \cite{deng_arcface_2019, he_deep_2016, roodschild_new_2020, hochreiter_recurrent_1998, kolbusz_study_2017, kajla_additive_2020, wang_loss_2020} is used instead of the cross-entropy loss function to obtain the best features in the face recognition problem. This is necessary when the input images have only one color, which makes it extremely difficult to identify the identity differences among the people in the photos. This model is trained with Glint 360k data \cite{an_partial_2021} consisting of about 170 million images of 360 thousand people from 4 different continents. The study uses the knowledge learned from this model to enrich the results of the problem. 
\item  Linear Mapping ($\mathcal{M}$): At the end of the model, an additional block of three fully-connected layers with the ReLU activation function is used to process the information and adjust its size to fit the Decoder components when the information is passed on to them. The pre-trained model will also be used to further train the data in the study.
\end{itemize}
\subsection{Geometry Decoder}
In this section, the study utilizes FLAME \cite{sanyal_learning_2019, cudeiro_capture_2019, li_learning_2017}, which is a well-known model for transferring feature information from 2D data vectors to anthropometric morphology of faces. This is a model that has been trained on 33,000 precisely aligned 3D faces. In MICA, this model is further trained on many other datasets including LYHM \cite{dai_statistical_2020}, FaceWarehouse \cite{cuturi_sinkhorn_2013, li_computations_2018, eckstein_computation_2021}, Stirling \cite{feng_evaluation_2018}. This extensive training allows the Decoder component to reconstruct important features such as nose shape, size, and face thickness. As previously discussed in the preceding section, the FLAME model is a well-known model for transferring features, and it is employed within the decoder part of the MICA model. By incorporating incomplete facial data into the transfer learning process, we can capitalize on the robust and well-optimized pre-trained MICA model. This approach is not only suitable for our study but also demonstrates its efficacy in harnessing the full potential of the pre-existing MICA model, thus enhancing the overall performance of the 3D face reconstruction process. 

In \cite{zielonka_towards_2022}, ArcFace architecture is expanded by a small mapping network $\mathcal{M}$ that maps the ArcFace features to their latent space, which can then be interpreted by geometry decoder:
\begin{align}
\mathbf{z} = \mathcal{M}(ArcFace(I)),
\end{align}
where $\mathbf{z} \in \mathbb{R}^{300}$ and $I$ is the RGB input image. Research by Zielonka et al. \cite{zielonka_towards_2022} used FLAME as a geometry decoder, which consists of a single linear layer: 
\begin{align}
\mathcal{G}_{3DMM}(\mathbf{z}) =  \mathbf{B}\times\mathbf{z} + \mathbf{A},
\end{align}
where $\mathbf{A} \in \mathbb{R}^{3N}$ is the geometry of the average human face and $\mathbf{B} \in \mathbb{R}^{3N \times 300}$ contains the principal components of the 3DMM and $N$ is the number of vertices of the output.  Then, the loss function of the MICA model is defined as follows
\begin{align}
\mathcal{L} = \sum_{(I, \mathcal{G}) \in \mathcal{D}} |k_{mask}(\mathcal{G}_{3DMM}(\mathbf{\mathcal{M}}(ArcFace(I)))  - \mathcal{G})|,
\end{align}
where $\mathcal{G}$ is the ground truth mesh, $D$ is the data set and $k_{mask}$ is a region dependent weight. The results of the transfer learning model with loss of training set and loss of validation set are shown in Fig. \ref{F9} and Fig. \ref{F10}.

\section{Recommendations from the study}
\label{sec:ssl}
Nevertheless, our primary objective does not solely rest on employing the transfer learning method for our new data; rather, we aspire to explore alternative approaches in order to identify the optimal solution for this research problem. Recognizing the potential of self-supervised learning as a promising and emerging technique, we have implemented this method in our study, which will be further discussed in the subsequent recommendations section.
\subsection{Self-supervised learning}
The self-supervised learning algorithm was adopted to improve the learning ability of the pre-trained Encoder component in our model. The objective is to enhance the model's ability to adapt to new data sets by first employing self-supervised learning to acquaint the model with the new data, after which supervised learning can be applied for fine-tuning. The idea is that if the model is trained on a large amount of data first, then the introduction of a small amount of new data into the model can cause a lot of weight changes, leading to poor performance. The self-supervised learning algorithm used in this model is Swapping Assignments Between Views (SwAV), which was developed by Caron et al. \cite{caron_unsupervised_2020} in 2020 and derived from the contrastive instance learning algorithm \cite{wu_unsupervised_2018}. The methods are based on typical clustering \cite{wu_unsupervised_2018, asano_self-labelling_2019} and this method is often used in the offline form which alternates between the assignment step into each model cluster. In this method, image features in the entire dataset are first clustered, and then assigned clusters are used to predict different views of the image during training. However, these clustering-based methods are not suitable for online learning models because they require many steps to compute the necessary image features for the clustering process. This method can be understood as comparing different views of the same image and comparing the cluster to which those images are assigned their features. More specifically, the research determines the code from an augmented version of the image sample and then predicts this code from other augmented versions of the same image. Given two image features $\mathbf{z}_t$ and $\mathbf{z}_s$ from two different augmented versions of the same image, we compute their code $\mathbf{q}_t$ and $\mathbf{q}_s$ by matching these attributes with a set of K prototypes $\{\mathbf{c}_1, \mathbf{c}_2, ...,\mathbf{c}_K\}$. After that, the study  builds the \textquotedblleft swapped\textquotedblright~ prediction step  by using the loss function as follows:
\begin{align}
L(\mathbf{z}_t, \mathbf{z}_s) = \ell(\mathbf{z}_t, \mathbf{q}_s) + \ell(\mathbf{z}_s, \mathbf{q}_t),
\label{eq1}
\end{align}
where the function $L(\mathbf{z}_t, \mathbf{z}_s)$ measures the fit between the attribute $\mathbf{z}_t$ and $\mathbf{q}_s$ the code.
\subsubsection{Online clustering}
Online clustering is a technique used in self-supervised learning of visual representations to group similar image patches or features into clusters. This technique is based on the idea that similar patches or features are likely to have similar representations in the learned visual space. In online clustering, the visual features of a large number of unlabeled images are extracted and clustered in an online manner as new images become available. The process starts with an empty cluster set and as new features are extracted from an image, they are assigned to one of the existing clusters or used to create a new cluster. Each $\mathbf{x}_n$ image is converted to an augmented view $\mathbf{x}_{nt}$ by applying the transformation $t$ derived from the set $\mathcal{T}$ of transformations. The augmented view is then mapped as a vector using a non-linear mapping function from $f_\theta$ to $\mathbf{x}_{nt}$. This feature is then projected onto the unit sphere, i.e., $\mathbf{z}_{nt}= f_\theta(\mathbf{x}_{nt})/||f_\theta(\mathbf{x}_{nt})||_2$. After that, the study computes the code $\mathbf{q}_{nt}$ from this attribute by mapping $\mathbf{q}_{nt}$) to a set of $K$ trainable prototypes vector $\{\mathbf{c}_1, \mathbf{c}_2, ...,\mathbf{c}_K\}$. The study is represented by a $\mathbf{C}$ matrix with columns $\{\mathbf{c}_1, \mathbf{c}_2, ...,\mathbf{c}_K\}$.
\subsubsection{Swapped prediction problem}
The loss function in Eq. (\ref{eq1}) has two terms that establish a \textquotedblleft swapped\textquotedblright~ prediction problem by predicting the code  from the attributes  and  from. Each term represents the cross-entropy loss function between the code and the probability obtained from using the softmax function that contains the dot product between  and all the prototypes in, i.e.,
\begin{align}
\ell\left(\mathbf{z}_t, \mathbf{q}_s\right)=-\sum_k \mathbf{q}_s^{(k)} \log \mathbf{p}_t^{(k)}, \quad \text { where } \quad \mathbf{p}_t^{(k)}=\frac{\exp \left(\frac{1}{\tau} \mathbf{z}_t^{\top} \mathbf{c}_k\right)}{\sum_{k^{\prime}} \exp \left(\frac{1}{\tau} \mathbf{z}_t^{\top} \mathbf{c}_{k^{\prime}}\right)},
\end{align}
where $\tau$ is the temperature parameter. Using this loss function for all images and pairs of augmentations yields a general loss function for the swapped prediction problem:
\begin{align}
-\frac{1}{N} \sum_{n=1}^N \sum_{s, t \sim \mathcal{T}}\left[\frac{1}{\tau} \mathbf{z}_{n t}^{\top} \mathbf{C} \mathbf{q}_{n s}+\frac{1}{\tau} \mathbf{z}_{n s}^{\top} \mathbf{C} \mathbf{q}_{n t}-\log \sum_{k=1}^K \exp \left(\frac{\mathbf{z}_{n t}^{\top} \mathbf{c}_k}{\tau}\right)-\log \sum_{k=1}^K \exp \left(\frac{\mathbf{z}_{n s}^{\top} \mathbf{c}_k}{\tau}\right)\right].
\end{align}
This loss function is minimized based on two $\mathbf{C}$ prototypes parameters and the  parameter of the image encoder $f_\theta$, which is used to output the $(\mathbf{z}_{nt})_{n,t}$ attribute.
\subsection{Computing codes online}
To enable the study to be used online instead of just offline, the study computes the codes by using only the image attributes within a batch to accommodate the usage of $\mathbf{C}$ prototypes across various batches, SwaV employed a clustering approach to group multiple versions of the prototypes. We calculated the codes by using the prototypes $\mathbf{C}$ in such a way that all samples in a given batch of images are equally partitioned among the prototypes. This even distribution makes each of the codes for different images in the batch unique in order to avoid trivial solutions where every image has the same code. Given the feature vector $\mathbf{B}$ consisting of $\mathbf{Z} = \mathbf{z}_1, \mathbf{z}_2, ..., \mathbf{z}_B$, this paper is interested in mapping them to the prototypes $\mathbf{C} = \mathbf{c}_1, \mathbf{c}_2, ..., \mathbf{c}_K$. The results represent these mappings as $\mathbf{Q} = \mathbf{q}_1, \mathbf{q}_2, ..., \mathbf{q}_K$ and optimizes $\mathbf{Q}$ to maximize the similarity between the properties and the prototypes:
\begin{align}
\max _{\mathbf{Q} \in \mathcal{Q}} \operatorname{Tr}\left(\mathbf{Q}^{\top} \mathbf{C}^{\top} \mathbf{Z}\right)+\varepsilon H(\mathbf{Q}),
\label{prob}
\end{align}
where $H$ is the entropy function, $H(\mathbf{Q})=-\sum_{i j} \mathbf{Q}_{i j} \log \mathbf{Q}_{i j}$ and $\varepsilon$ is the parameter that controls the smoothness of the mapping. It is observed that a strong entropy regularization (that means using high $\varepsilon$) can often lead to a trivial solution, where all templates collapse into a unique representation and all prototypes have the same property. Therefore, the parameter $\varepsilon$ should be fixed at a small value. To ensure equal partitioning, Asano et al. \cite{asano_self-labelling_2019} constrained the matrix that belonged to the transportation polytope. To address this issue, the present study proposes an adaptation of the solution proposed by \cite{caron_unsupervised_2020}, which involves constraining the transportation polytope to the minibatch when working with minibatches as follows:
\begin{align}
\mathcal{Q}=\left\{\mathbf{Q} \in \mathbb{R}_{+}^{K \times B} \mid \mathbf{Q} \mathbf{1}_B=\frac{1}{K} \mathbf{1}_K, \mathbf{Q}^{\top} \mathbf{1}_K=\frac{1}{B} \mathbf{1}_B\right\},
\label{eq_Q}
\end{align}
where $\mathbf{1}_K$ is a vector of ones with $K$ dimensions. These constraints ensure that on average each prototype selects at least $B/K$ times in a batch. When a continuous solution for $\mathbf{Q}^*$ in Eq. (\ref{eq_Q}) is found, the discrete code can be calculated using the rounding process. In an online setting, we used mini-batches with discrete codes that performed worse than continuous codes. These soft codes $\mathbf{Q}^*$ are the solution of Prob. (\ref{prob}) over the set $\mathcal{Q}$ is written as
\begin{align}
\mathbf{Q}^*=\operatorname{Diag}(\mathbf{u}) \exp \left(\frac{\mathbf{C}^{\top} \mathbf{Z}}{\varepsilon}\right) \operatorname{Diag}(\mathbf{v}),
\end{align}
in which, $\mathbf{u}$ and $\mathbf{v}$ and are the renormalization vectors in $\mathbb{R}^K$ and $\mathbb{R}^B$ respectively. The vector renormalization is calculated using a small number of matrix multiplications along with the iterative Sinkhorn-Knopp algorithm \cite{h_f_loshchilov_decoupled_2019}.
\subsection{Post-processing: Filling Extraction}
To achieve completion of the wound healing process and effectively aid in concealing facial imperfections, we have used to employ the filling extraction technique, which was initially developed and discussed in one of our prior publications \cite{nguyen_3d_2023}. This innovative approach is specifically tailored to analyze the wound-affected area on the face, isolate it, and extract it separately for subsequent utilization in 3D printing applications.

\begin{figure}[!h]
\centering
\includegraphics[scale=0.45]{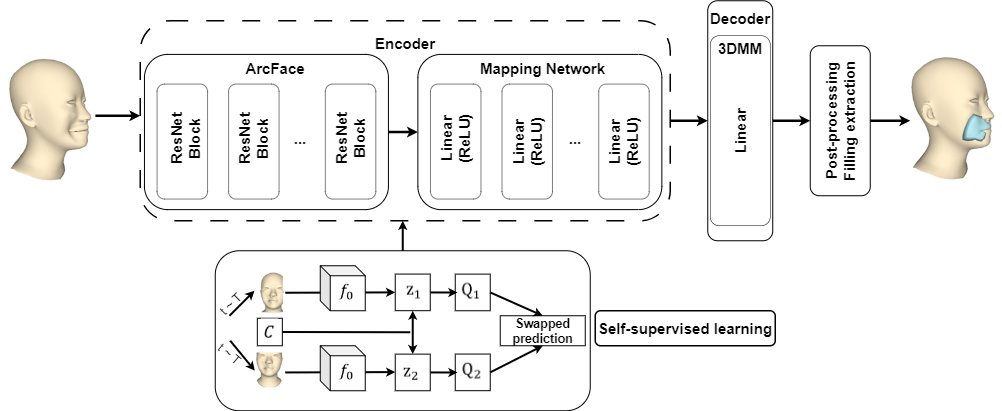}
\caption{A model for supervised learning.}
\label{F7}
\end{figure}

By employing this technique, our final printable product is an amalgamation of the original, imperfect face and the separately extracted wound portion. Notably, the extracted part is displayed in a distinct color to differentiate it from the rest of the facial structure, as clearly demonstrated in Fig. \ref{F7}. The combination of these elements allows for a more comprehensive understanding of the wound area and its impact on the overall facial reconstruction process.
\section{Experiments}
\subsection{Building Dataset}
In order to effectively implement the transfer learning and self-supervised learning processes within the MICA model, we have undertaken a comprehensive data preprocessing approach, which consists of two primary steps. Each step has been carefully designed to ensure the optimization of our model's performance and applicability in the context of reconstructing incomplete facial structures.

The first step focuses on preparing the input data in a 2D format. This decision is grounded in two main considerations. Firstly, utilizing a 2D input format is consistent with the MICA model's architectural requirements, thus allowing for seamless integration of the data. Secondly, the adoption of 2D input data presents an alternative method for facial reconstruction, which significantly simplifies the process for medical professionals. As a result, physicians are no longer required to provide 3D MRI scans; instead, they can rely on more readily available 2D images. This adaptation seeks to create a more conducive environment that enables healthcare practitioners to efficiently engage with our model, ultimately delivering enhanced value and outcomes for patients. Additionally, patients can preview their facial appearance after treatment, providing them with a clear understanding of the expected results. The second step of the data preprocessing procedure involves employing augmentation techniques within the mathematical space to standardize the model's output. This standardization serves as the Ground Truth for the MICA model to calculate the loss value, ensuring that the reconstructed facial structures are accurate and reliable. By integrating these augmentation techniques, we aim to improve the model's overall performance and facilitate a more precise reconstruction of facial defects.
The specific details of the input and output parameters are described in greater detail below:
\begin{itemize}
    \item The 2D input format is obtained by capturing images from all 3D data with varying lighting directions. The 2D images have a resolution of 1024x1024 pixels and are captured from 3687 3D samples within our dataset, ensuring that the angle of capture clearly reveals any scarring present.
    \item The MICA model's output, with 3d format, is modified to match with the Ground Truth. By default, MICA generates eyeballs for the output mesh but the Ground Truth mesh of the Cir3D-FaIR dataset has no eyeballs, so we have conducted eyeball removal procedures for the entire set of MICA model's outputs. Another problem is that 3D printing needs a watertight mesh to have a good processing step, but the output mesh of the MICA model has a hole that makes the mesh non-watertight, therefore the hole-fixing procedures need to be used. The computational techniques of eyeball removal and hole-fixing procedures introduced in our previous research \cite{nguyen_3d_2023} and employed in this space have been encapsulated within the CirMesh library and made available in Python by us, with the easy installation using the command {\color{red}{"pip install cirmesh"}}. After that, the MICA model's output and the Ground Truth mesh could compare with each other.
\end{itemize}

To prepare for the training process, the research divides the data into a 6:2:2 ratio for the training, validation, and test sets, respectively. During the data splitting procedure, the study ensures that images of different individuals are allocated to each dataset, preventing the exchange of facial information among the three sets.

\subsection{Determine the set of parameters in the research model}
Runtime: The experiments were conducted on google colab pro. The proposed model requires computational cost and runtime. The first epoch takes about 30 minutes for the training and validation process, while other epochs take only about 90 seconds to complete as the data is already stored.

\begin{figure}[!h]
\centering
\includegraphics[scale=0.5]{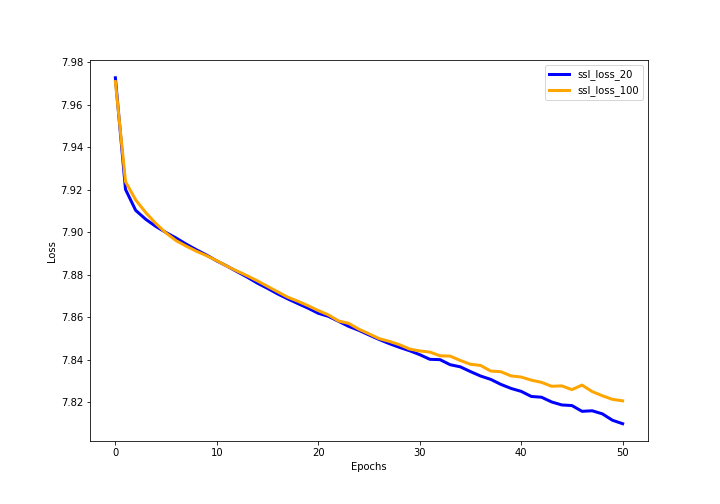}
\caption{Loss function value corresponding to each Epochs}
\label{F8}
\end{figure}

Hyperparameters: In all experiments, this research fixed the version of hyperparameters. The optimal algorithm is Adam \cite{h_f_loshchilov_decoupled_2019} whose initial learning rate value is $1\times10^{-3}$, which is kept constant throughout the training. At the same time, a batch size of 50 images is used for each epoch. Finally, the experimental process is conducted for 50 epochs for each case as shown in Fig. \ref{F8}.

The results in Fig. \ref{F8} indicate the change in the loss function value for each epoch in two cases: 20\% data and 100\% data. Both cases share a similar pattern of loss function change for the first 30 epochs. The case with 100\% data shows faster convergence in the following 20 epochs. This suggests that the present model is highly stable.
\subsection{Experiment settings}
\begin{figure}[!h]
\centering
\includegraphics[scale=1]{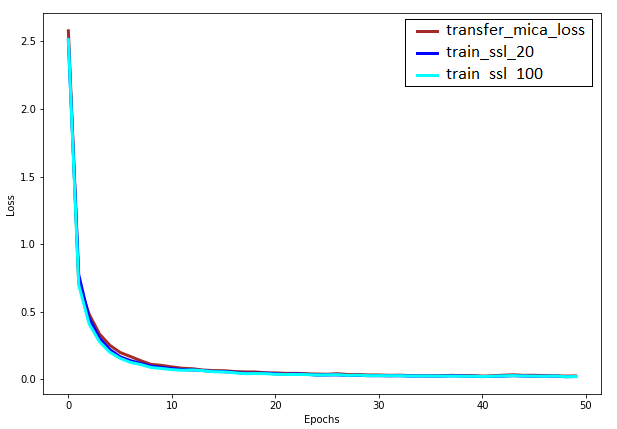}
\caption{Loss of training set}
\label{F9}
\end{figure}

The study conducted a total of 5 experiments. In the first experiment, the study used the pre-trained model of MICA \cite{zielonka_towards_2022} with its enriched data. In the remaining cases, the study applied the self-supervised learning technique as mentioned in Section \ref{sec:ssl}, to train the Encoder component before proceeding to the supervised learning process. The study conducted 50 epochs with self-supervised learning in two main cases. As discussed earlier, the study initially used a sample of 20\% of the training data and then used a larger sample to evaluate the loss of the proposed models. Fig. \ref{F9} shows the loss of self-supervised algorithm training when sampling 20\% and 100\% of the data. When approaching the 50th epoch, the loss function of the method with 20\% of the data sample performed better than that with 100\% of the data sample. This can be explained by the fact that when dealing with larger amounts of data, self-study may have some difficulties in adapting to the new data.

\begin{figure}[!h]
\centering
\includegraphics[scale=1]{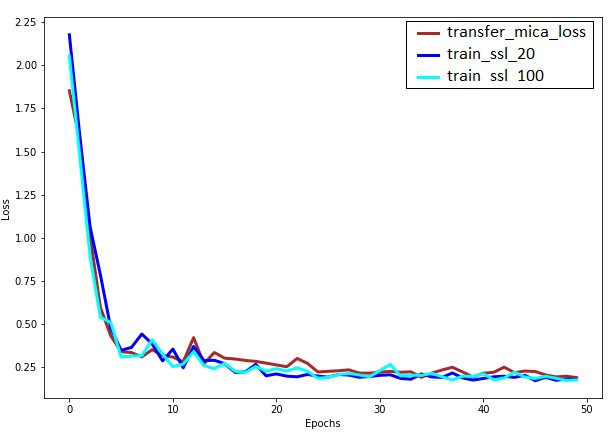}
\caption{Loss of validation set}
\label{F10}
\end{figure}

During the training process, the results in Fig. \ref{F10} show that the loss function path of the training set of all three cases is stable and all values go below 0.5 at the 10th epoch. After that, the loss line decreased slightly and approached near zero at the 50th epoch. In all cases, the loss lines are not significantly different from each other. For the validation set, the loss lines seem to be volatile at the 10th epoch, and then become stable again as they move toward the 50th epoch.

\begin{table}
\renewcommand{\arraystretch}{1.5}
\centering
\caption{Geometric distance Function values of the outcomes }
\label{Tab:1}
\begin{tabular}{|l|c|}
\hline Method & Mean distance \\
\hline Transfer Learning & 0.3642 \\
\hline Self-supervised Learning with 20\%  data  & 0.3353 \\
\hline Self-supervised Learning with 100\%  data  & 0.3247 \\
\hline
\end{tabular}
\end{table}

In our prior research \cite{nguyen_3d_2023}, the outcomes of the developed model were assessed using the Geometric Loss Function. To evaluate the current models in this study, we computed the error for the transfer learning model and two instances of self-supervised learning at both 20\% and 100\% data utilization, employing the Geometric Loss Function as demonstrated in the accompanying Table \ref{Tab:1}. Initially, when utilizing the transfer learning method, the observed loss of 0.36 signifies a satisfactory reconstruction of facial injuries. By applying the self-supervised learning method from \cite{caron_unsupervised_2020} on 20\% of the training data, the loss function improves to 0.3, exhibiting an enhancement of approximately 0.06. Subsequently, increasing the training data proportion to 100\% and implementing the self-supervised method results in a further improvement of the loss function by roughly 0.01, compared to the previous method that employs 20\%.

These findings suggest that employing a limited amount of data in the pre-trained model, using the self-supervised learning technique is a feasible approach that offers superior efficiency compared to the transfer learning method. Notably, our datasets comprise only 3687 images, accounting for less than one-tenth of the data in the purely pre-trained MICA model. As a result, our approach effectively leverages the capabilities of the trained MICA model while remaining applicable to the current research context.
\begin{figure}[!h]
\centering
\includegraphics[scale=0.6]{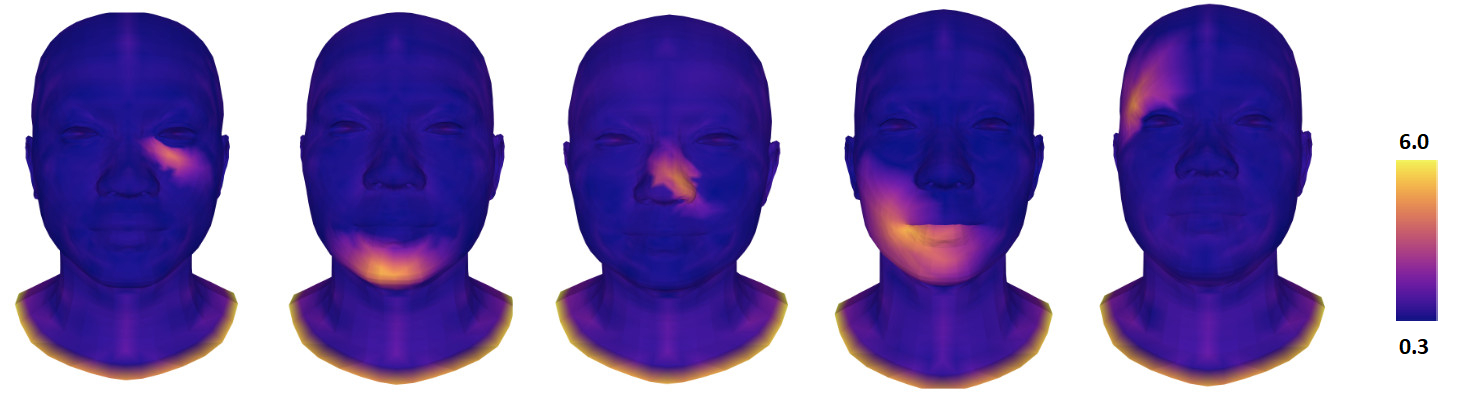}
\caption{Deviation between results and reality}
\label{F11}
\end{figure}

\begin{figure}[!h]
\centering
\includegraphics[scale=0.7]{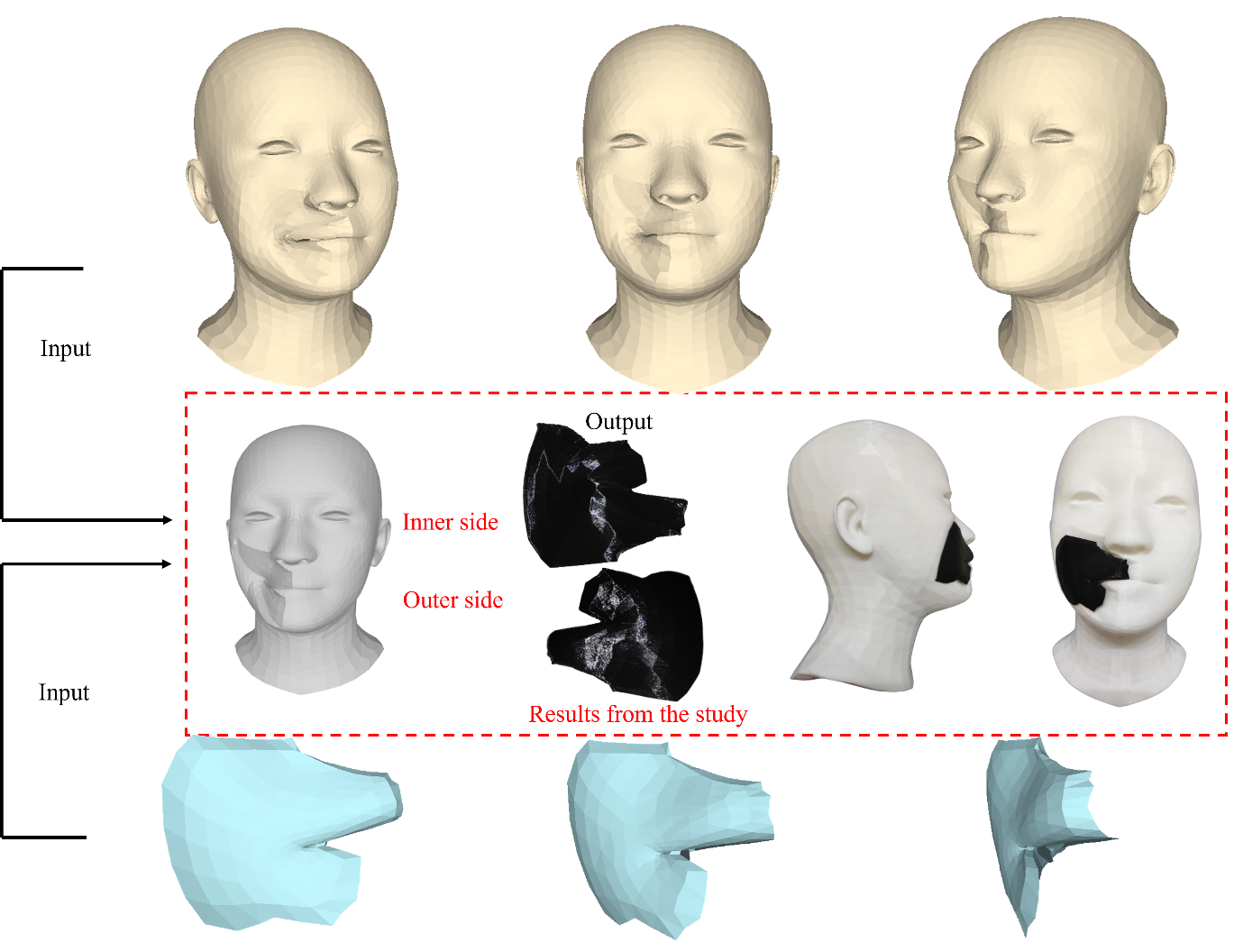}
\caption{Results from the study using 3D printed models}
\label{F12}
\end{figure}

Fig. \ref{F11} shows some examples illustrating the model’s ability to reconstruct the structure of human facial wounds and the deviation between the output and the reality. Brighter regions signify larger errors, which can be identified as wounds. Although it is inevitable that some errors will occur while reconstructing injured areas, the proposed model successfully reconstructed the majority of the face, leaving virtually no unaffected regions. These findings suggest that the model is proficient in addressing the challenges of reconstructing defects stemming from wounds, even when solely relying on 2D input. The results of this study have been applied to the creation of experimental products using 3D printing technology, as shown in Fig. \ref{F12}, to be utilized in laboratory experiments. The results of this study reveal significant potential for practical applications, especially in creating models that mask flaws in the human face.

\section{Conclusions}
This paper illustrates the advantages of utilizing a pre-trained model for 3D Face Reconstruction, as it significantly benefits the underlying problem by enabling the proposed approach to capitalize on the capabilities of existing models. The study presents several noteworthy findings:
\begin{itemize}
    \item All data added can be incorporated without requiring a complete retraining process from scratch. This results in an efficient method that does not demand an extensive amount of training data. Such efficiency is achievable through the strategic integration of pre-trained models and self-supervised learning, which boosts performance without adding complexity to the model. The results of this study demonstrate the feasibility of reconstructing incomplete faces using only 2D images, providing surgeons and orthopedic trauma specialists with additional data and improved procedures to better support their patients.
    \item The training time has been significantly reduced from a duration of several days to a matter of mere hours, facilitating the extraction of facial features from incomplete faces. Moreover, this reduction allows for the implementation of the training process on platforms like Google Colab without necessitating a high-performance computer. 
    \item Although the obtained results exhibit a relatively wide range of errors, from 0.3 to 6 as shown Fig. \ref{F11}, which is larger than the error margins observed in our previous research, the average error remains at a manageable level of 0.32, indicating that the overall effect is not significantly compromised. This can be easily understood, as the method attempts to extract information from a 2D format, which inherently lacks depth information compared to the 3D format.
\end{itemize}

\section*{Acknowledgments}
We would like to thank Vietnam Institute for
Advanced Study in Mathematics (VIASM) for hospitality during our visit in 2023, when we
started to work on this paper. We would like to thank MSc Thanh Q. Nguyen who is a researcher at CIRTECH Institute, HUTECH University, Ho Chi Minh City, Vietnam for supporting us in the completion of this study. It is important to highlight that the introduction of this article underwent refinement to enhance its readability, using a Large Language Model (LLM) called ChatGPT from OpenAI, based on our initial paper. Authors are accountable for the originality, validity, and integrity of the content of our submissions. The use of this AI-assisted technology is permitted by the 
\href{https://newsroom.taylorandfrancisgroup.com/taylor-francis-clarifies-the-responsible-use-of-ai-tools-in-academic-content-creation}{publisher's policy}.

\bibliographystyle{unsrt}  
\bibliography{3D_FaIR_SSL}

\end{document}